# Detect Toxic Content to Improve Online Conversations

Deepshi Mediratta (300086013)
School of Electrical Engineering and Computer Science (EECS)
University of Ottawa
Ottawa, Canada
dmedi018@uottawa.ca

Nikhil Oswal (300074118)
School of Electrical Engineering and Computer Science (EECS)
University of Ottawa
Ottawa, Canada
noswa023@uottawa.ca

***Abstract*—Social media is filled with toxic content. The aim of this paper is to build a model that can detect insincere questions. We use the 'Quora Insincere Questions Classification' dataset for our analysis. The dataset is composed of sincere and insincere question, with majority of sincere questions. The dataset is processed and analyzed using Python and its libraries such as sklearn, numpy, pandas, keras etc. The dataset is converted to vector form using word embeddings such as GloVe, Wiki-news and TF-IDF. The imbalance in the dataset is handled by resampling techniques. We train and compare various machine learning and deep learning models to come up with the best results. Models discussed include SVM, Naïve Bayes, GRU and LSTM.**

## I. Introduction

With so many social media platforms existing on the internet today, the amount of content created online on a daily basis is growing at an ever-increasing rate as more of our world becomes digitized [1]. Social media websites have been focusing on encouraging people to participate in content generation, but have paid less attention to the content itself.

Identifying and eliminating 'toxic' content has become a problem for any major website today. By the time a user reports any harmful content and any action is taken by the website, the content could have done a ton of damage already. Social media websites such as YouTube have recently started taking serious actions in order to remove problematic content from their websites [2] [3]. Facebook rolled out a 'protective detection' system designed to flag posts that come from people threatening suicide or self- harm and posts that are aggressive towards others [4].

Quora is a website that encourages people to ask questions and learn from each other [5]. However, they also face a challenge of removing insincere questions - questions that are founded on false premises, or that intend to make a statement rather than look for helpful answers. They want to tackle this problem so that their users can feel safe sharing their knowledge.

This study aims to harness machine learning and deep learning techniques to screen content before it is posted online on various social media groups and websites. In this work, we examine the dataset provided by Kaggle website [6] regarding Quora Insincere Question Classification, and employ machine learning and deep learning techniques to discover toxic questions.

The paper is organized as follows. In Part II, we talk about some of the related work. Part III focuses in the problem statement, description of the dataset and the exploratory data analysis. Part IV describes the methodology used to solve the problem- how the data was prepared and represented, what word embeddings were used, how the imbalance in the dataset was handled. We also explain the models used, how these models work, how we implemented the model. In Part V, we eplain how the dataset was divided for training and test purpose and what all evaluation criteria was used to evaluate the model's performance. In Part VI, we present the results obtained using various word embedding and classifiers. In Part VII, we discuss these results. Part VIII provides a summary of the work done.

## II. Related Work

Detecting inappropriate and negative content is a highly relevant problem today. A lot of work has been done in this area using machine learning and deep learning models. Most of the research focuses on classifying large texts such as movie review, blogs posted online etc. The authors of the paper [7], employ Convolutional Neural Networks to discover toxic comments in a large pool of documents provided by Kaggle regardin Wikipedia's talk page edits. Researchers have performed sentiment analysis on various social media websites such as IMDB [8], Wikipedia [7] etc., to analysis the emotions and intent of the users. Microblogging websites such as Twitter [9] [10] are also analyzed in various papers. The various models used for sentiment classification are LSTM, Naïve Bayes and SVM. The work done in [11] is based on variations of BERT model on the Quora insincere Question Classification dataset. We show that we can produce comparable results on the same dataset using basic machine learning models such as Support Vector Machine and Naïve Bayes and deep learning models such as GRU and LSTM.

## III. QUORA INSINCERE QUESTIONS CLASSIFICATION

### *A. Data description*

We have used an open dataset by Quora which have over one million rows [6]. The dataset consists of questions in English that users have posted online on Quora. The target values are 0 or 1, corresponding to weather the question should be classified as 'sincere' or 'insincere respectively.

According to the Kaggle competition, an insincere question is defined as a question intended to make a statement rather than look for helpful answers. Some characteristics that can signify that a question is insincere:

- Has a non-neutral and exaggerated tone to underscore a point or imply a statement about a group of people.
- Is disparaging or inflammatory, such as attacks or insults against a specific person or a group of people, based on an outlandish premise about a group of people, or disparages against a characteristic that is not fixable and not measureable.
- Isn't grounded in reality, is based on false information, or contains absurd assumptions.
- Uses sexual content (incest, bestiality, pedophilia) for shock value, and not to seek genuine answers

Some examples of insincere questions include:

- Why do Chinese hate Donald Trump?
- Do Americans that travel to Iran have a mental illness?
- How do I train my dogs to kill raccoons?
- How crazy are religious people?
- Why is the French army so bad?

The insincere questions in the text consists of a wide range of topics that are challenging to detect as 'sincere' or 'insincere'. If we find some words such as ugly, kill, crazy, mental etc. in a sentence, it is easy to identify them as insincere. However other topics such as Republican, Democracy etc. are difficult to classify as insincere as these words could be used in sincere questions also.

File descriptions: train.csv training set of size 1.31m * 3

Data fields:

- qid: unique question identifier.
- question_text: Quora question text.
- target: a question labeled 'insincere' has a value of 1, otherwise 0.

### *B. Exploratory Data Analysis*

The dataset contains 1,306,122 set of sincere and insincere questions. The size of the dataset may impose additional challenges of running into memory errors and could also lead to high processing times.

If we have a look at the target variable distribution, we observed that there is a high imbalance in the two classes. We only have approximately 6% of data that is classified as insincere. If we train a machine learning model without handling the imbalance, the model could be highly biased towards the sincere class. We need to address this imbalance in the dataset, before training our models, by using resampling techniques and data augmentation.

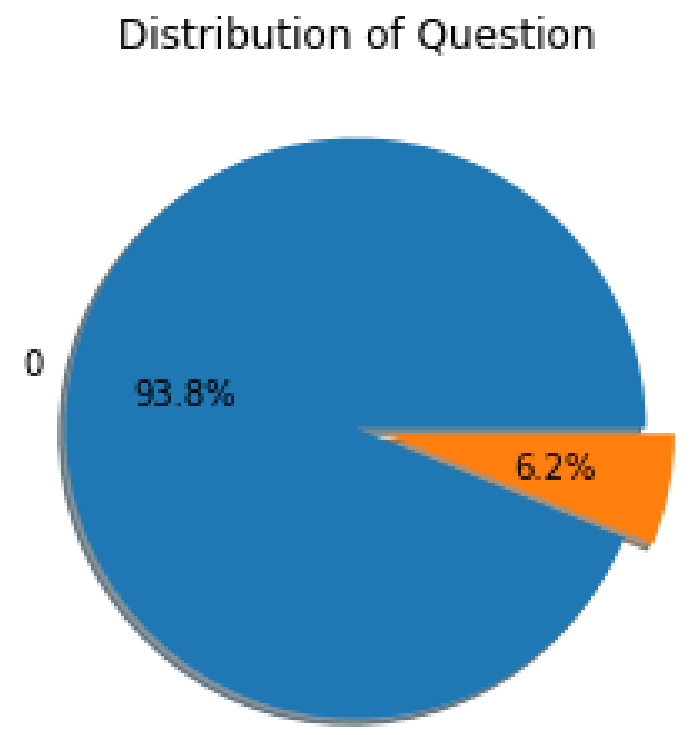

Fig. 1. Distribution of the target variable showing imbalance in the dataset.

To analyze the word frequency distribution in the insincere questions, we first need to clean the data and remove words which are not meaningful. We applied some data cleaning techniques such as- tokenizing sentence to words, converting words to lowercase, removing punctuations, removing stop words, removing words whose length is less than 4. When we plot the Word frequency distribution chart (Figure. 1) we observed that the most frequently used words in the insincere questions were – people, Trump, women, think, Muslims, Quora, India, white, Indian, Americans etc.

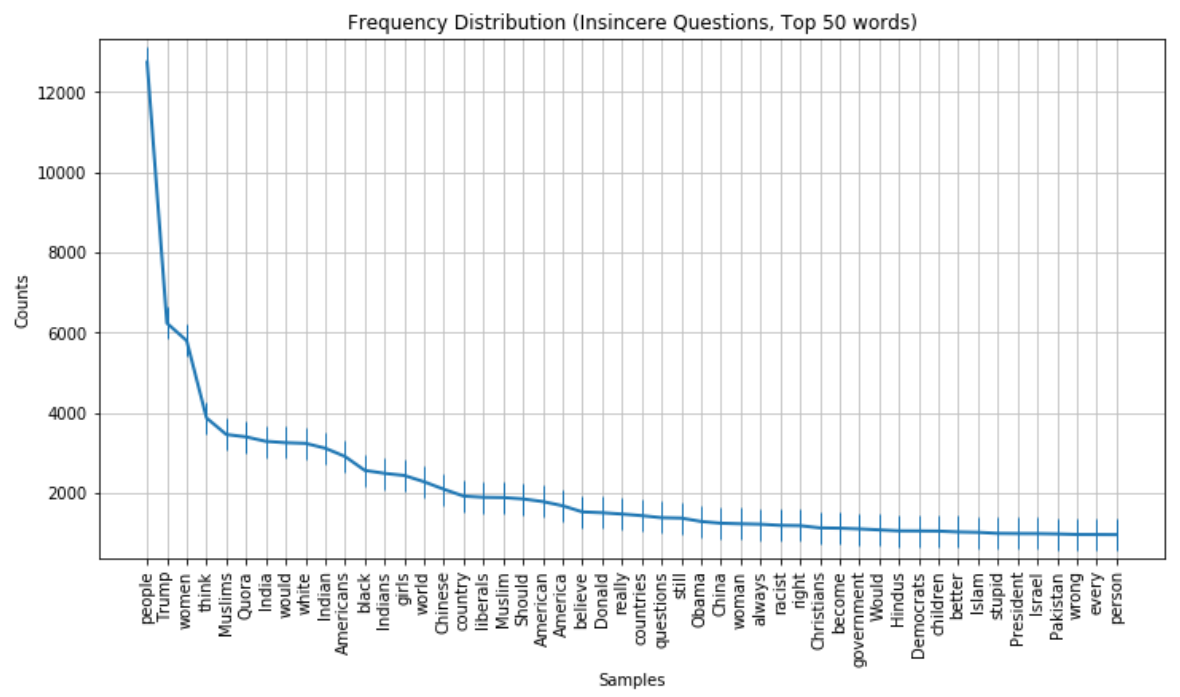

Fig. 2. Frequency Distribution of words in insincere questions.

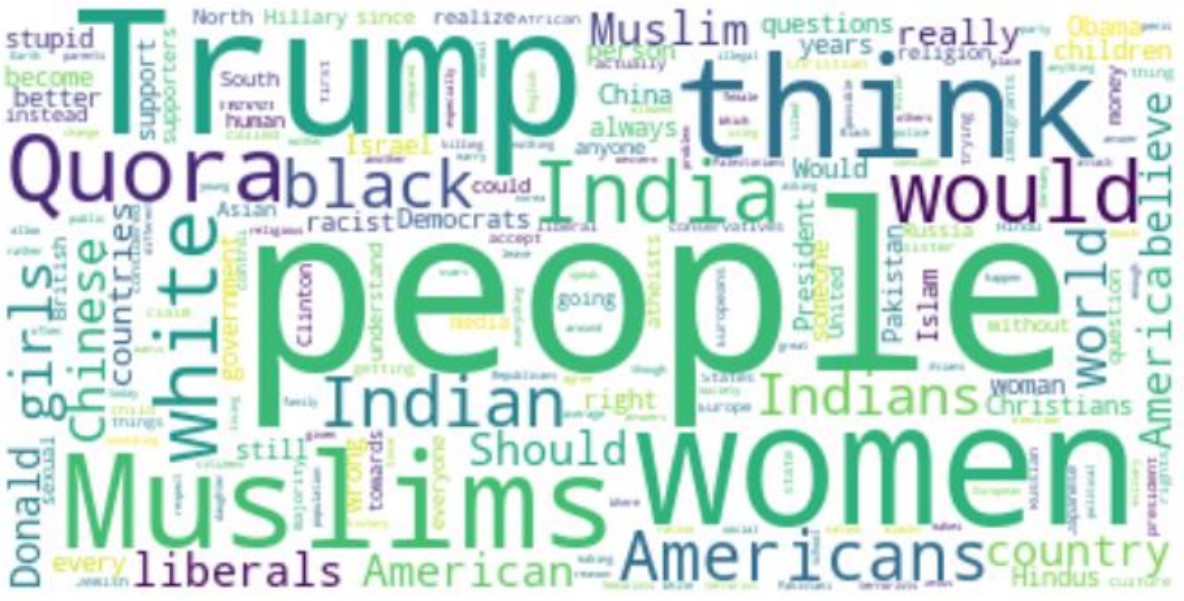

Fig. 3. Word Cloud of the most frequent words in insincere questions.

## IV. METHODOLOGY

### A. Preparing the data

The dataset was in csv format with 1,306,122 samples of questions and 3 columns. We converted them into pandas data frame for easy data manipulation. The target values are already in binary format so we don't need to manipulate them, they can be fed as is in the models. We extract the values of the column 'question_text' in a variable X. The questions are in text format. Text format is not compatible with machine learning and deep learning models, so we will have to convert them to suitable formats.

### B. Word Embeddings

We map the words in the questions to their respective vector formats as machine learning and deep learning models are not able to process strings or plain text in their raw format.

For text preprocessing, we have used The Tokenizer class available in python's keras library. This class allows to vectorize a text corpus, by turning each text into a vector.

First, we use the 'text_to_sequences' method to convert the text to vectors. The top most frequent words are taken into account. Only the words known by the tokenizer will be taken into account. Next, we transform the vector sequences into a 2D numpy array of shape (number of samples X 100) using the pad_sequences method.

The word embedding that we have worked with for this dataset are:

1. **Glove**: GloVe stands for global vectors for word representation [12]. We used GloVe embedding's which is based on factorizing a matrix of word co-occurrence statistics. We experimented on all dimensions of glove vectors and decided to work with 100-dimensional GloVe vectors as they provided the best results for our task

2. **Keras Embedding Layer**: Keras offers an Embedding layer that can be used for neural networks on text data [13]. The Embedding layer is defined as the first hidden layer of a network with three arguments – input dimension, output dimension and input length.

3. **Wiki-news**: This embedding consists 1 million word vectors trained on Wikipedia 2017, UMBC webbase corpus and statmt.org news dataset [14]. We observed that the most frequent words used in the insincere questions were related to the recent hot topics in news. So using wiki-news embeddings would be helpful.

4. **TF-IDF**: In a large text, we often encounter texts such as 'a', 'is', 'an' etc which do not provide much meaning. These words are called stopwords. TF-IDF [15] approach addresses the stopwords issue by using statistical quantity- *tf( term, document).idf(term)* The first part is tf which stands for term frequenct and the second part is idf which stands for inverse document frequency. TF-IDF converts a collection of raw documents to a matrix of TF-IDF features.

$$tf(term, document) = \frac{n_i}{\sum_{k=1}^{V} n_k}$$

$$idf(term) = log\frac{N}{n_t}$$

### C. Data Representation

Using word embeddings, we transformed our text features in form of vectors. We form tensors from these word vectors which can be fed in the machine learning and deep learning models. Sequences that are short are padded with value 0. We do not update the pre-trained word embeddings during training.

Example of encoding of a sentence to vectors using 100D GloVe:

Question: 'What would happen if every US citizen did not want to vote due to lack of interest?'

Padded Sequence: [0, 0, 0, 0, 0, 0, 0, 0, 0, 0, 0, 0, 0, 0, 0, 0, 0, 0, 0, 0, 0, 0, 0, 0, 0, 0, 0, 0, 0, 0, 0, 0, 0, 0, 0, 0, 0, 0, 0, 0, 0, 0, 0, 0, 0, 0, 0, 0, 0, 0, 0, 0, 0, 0, 0, 0, 0, 0, 0, 0, 0, 0, 0, 0, 0, 0, 0, 0, 0, 0, 0, 0, 0, 0, 0, 0, 0, 0, 0, 0, 0, 0, 0, 0, 0, 2, 35, 188, 20, 236, 92, 1346, 48, 44, 85, 5, 1280, 526, 5, 1541, 7, 810]

### D. Resampling Techniques

We observed that we have an imbalance in the dataset. The learning phase and the prediction of machine learning and deep learning algorithms can be affected by the problem of an imbalanced dataset. The decision function of the models might favor the class with the larger number of samples.

We need to down-sample the majority class, i.e., the sincere class, to equal the size of the minority class, i.e., the insincere class. This will help us balance the two classes in the dataset as well as decrease the processing time as the number of samples in the training data will also decrease.

We used the under-sampling techniques provided by imbalanced-learn [16].

**ClusterCentroids**: It is a method that under samples the majority class by replacing a cluster of majority samples by the cluster centroid of a KMeans algorithm. In centroid based majority under sampling technique, unimportant instances are removed among majority samples. The cluster centroid is found by finding the average feature vectors for all the features, over the data points belonging to the majority class in feature space. After the centroid is found, the majority class sample nearest to the centroid is considered to be the most important feature.

**RandomUnderSampler**: It is a fast and easy way to balance the data by randomly selecting a subset of data for the targeted class with or without replacement. It allows to bootstrap the data by setting the replacement parameter to true. The resampling with multiple classes is performed by considering independently each targeted class. It also allows data that contains strings.

For our experiments, we experimented with both ClusterCentroids and RandomUnderSampler but decided to work with RandomUnderSampler because it was fast and provided good results.

*E. The Models*

1. **SVM**: Support vector machine works by finding and constructing a hyperplane in N-dimensional space that separates the points between two classes, N being the number of features. The hyperplane is determined by finding a plane that has the maximum margin which is the distance between the data point of two classes. Points that fall on the side of the hyperplane can be attributed to different classes. We used the Support Vector Classifier provided by sklearn [17] for training and testing. The kernel type to be used in the algorithm is 'linear' ($x$, $x'$). The degree of the polynomial kernel function is chosen as 3. The gamma parameter is set to 'auto' which uses 1/ n_features.

2. **Naïve Bayes**: This model is based on Bayes' theorem with the naïve assumption of independence between each pair of features. If we need to classify the vector X = $x_1 \ldots x_n$ into m classes, $C_1 \ldots C_m$. we need to find the probability of each class given X. Then we can assign X the label of the class with highest probability. The probability is calculated using Bayes' theorem which is defined as:

$$P(C_i|X) = \frac{P(X|C_i)P(C_i)}{P(X)}$$

   We used the Naïve Bayes classifier for multinomial models provided by sklearn [18] with default parameters. We chose the multinomial Naïve Bayes classifier because is appropriate for text classification. For this classifier we have used TF-IDF fractional count representation of the features.

3. **GRU**: GRU are improved version of standard recurrent neural network. RNN have the issue of vanishing gradient problem, i.e., when they are learning weights using back propagation, the weights get out of bound and their values become very less. GRU uses update gate and reset gate to address the vanishing gradient problem. They can be trained to keep information from long ago and are able to remove information which is irrelevant to the prediction.

   We have used CuDNNGRU [19], which is a fast GRU implementation backed by CuDNN (a GPU-accelerated library of primitives for deep neural networks). It can be only run on a GPU, with the TensorFlow backend.

   Embeddings – 3 experiments. Without pretrained embeddings, using glove embeddings and using wiki news FastText Embeddings.

   Activation Function – Relu for the middle dense layer, Sigmoid for the last dense layer.

   Loss Function – binary cross entropy.

   Optimizer – Adam

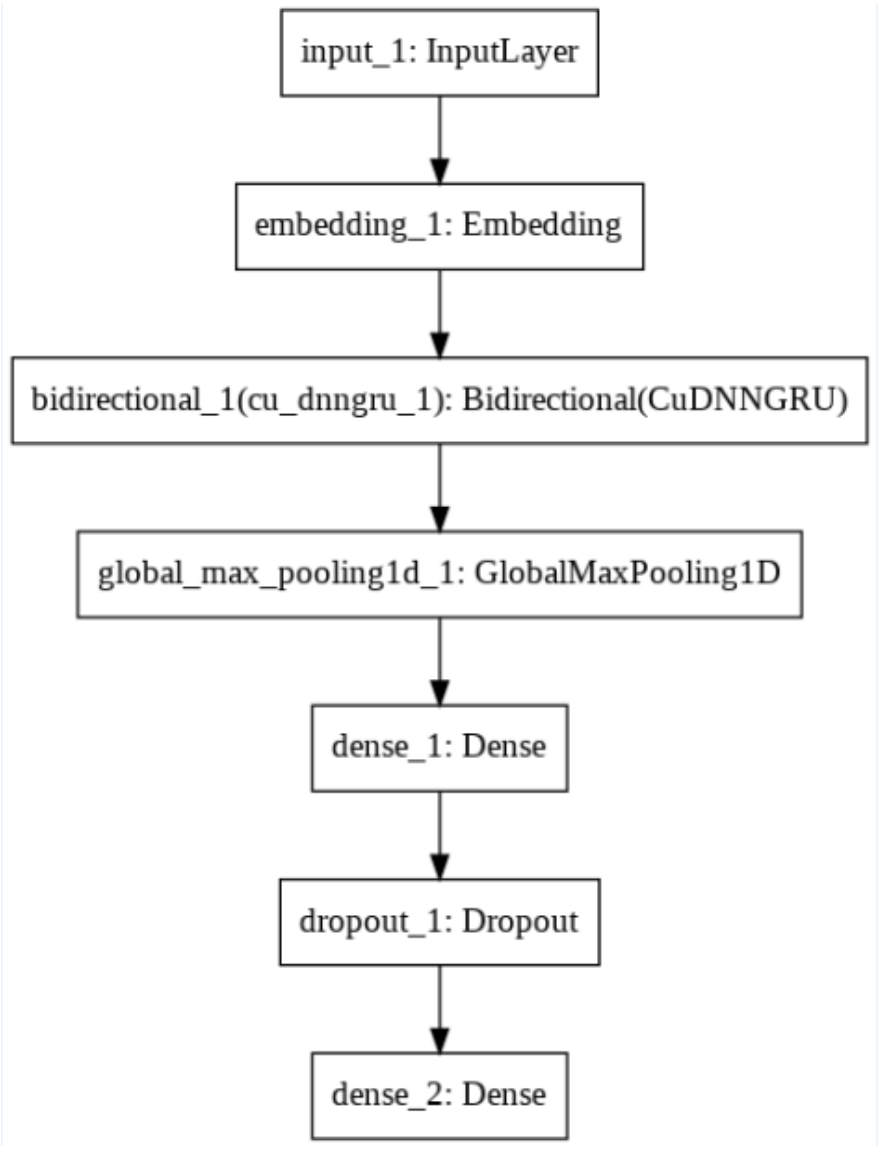

Fig. 4. Architecture of the Bidirectional CuDNNGRU model.

4. **LSTM**: Long short-term memory layer is a feed forward neural network. Vanilla RNN fail to understand the context behind an input. They are not able to recall some text that they saw long back to make predictions in the present. LSTM are able to choose ehat information should be remembered or which should be forgotten. They make use of a Forget gate, input gate and Output gate to do so.

   We have used LSTM implementation by keras [20]

   Embeddings – 3 experiments. Without pretrained embeddings, using glove embeddings and using wiki news FastText Embeddings.

   Activation Function – Relu for the middle dense layer, Sigmoid for the last dense layer.

   Loss Function – binary cross entropy.

   Optimizer – Adam

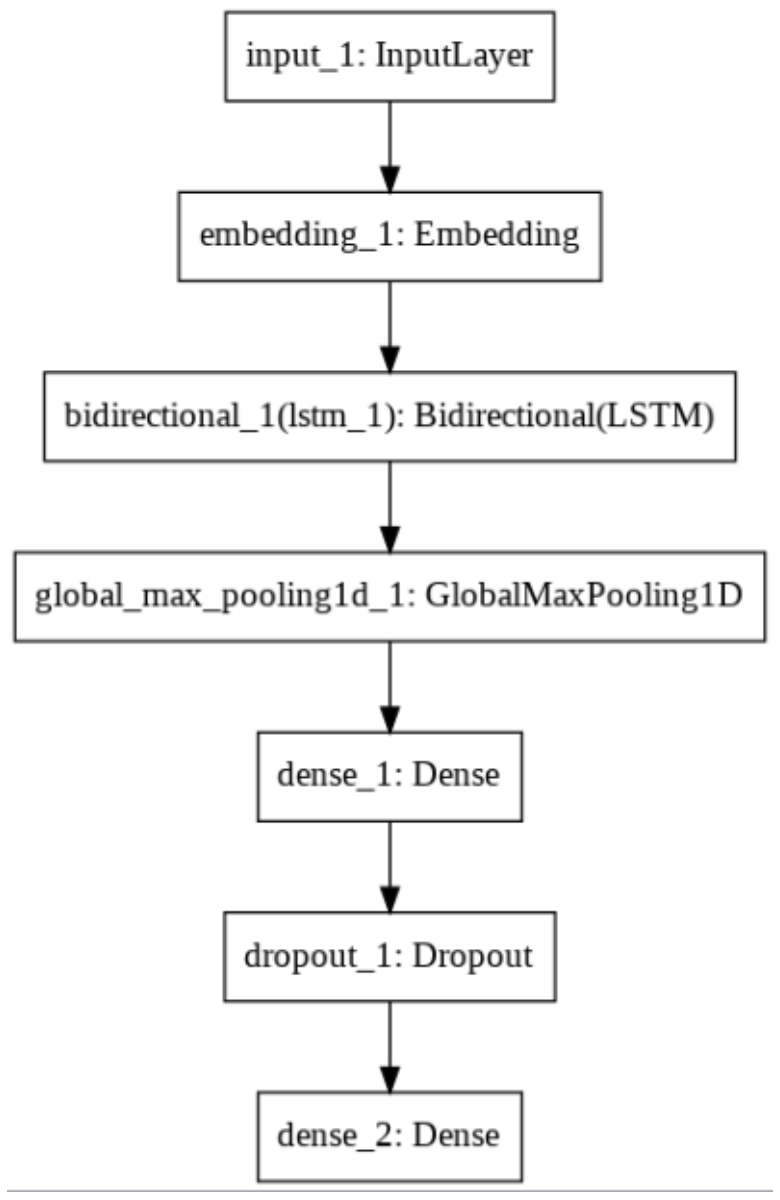

Fig. 5. Architecture of the Bidirectional LSTM model.

## V. Experiments

### A. *Training and Test Datasets*

We split the dataset into training and test data. The training set contains a known output and is used for learning. The test dataset is used to test the model's prediction. We do this using the method train_test_split provided by the Scikit-learn library. We split the data set in 80:20 ratio, 80% for the training and 20% for testing.

### B. *Evaluation Metrics*

**Accuracy:** The ratio of correct predictions over the total predictions.

$$Accuracy = \frac{TP + TN}{TP + TN + FP + FN}$$

**Precision**: The number of true positives divided by all positive predictions. It is a measure of a classifier's exactness. It tells us how often the classifier is correct when it predicts positive. Low precision means that there is a high number of false positives.

$$precision = \frac{TP}{TP + FP}$$

**Recall**: The number of true positives divided by the number of positive values in the test data. It is also known as Sensitivity or the True Positive Rate. It is a measure of a classifier's completeness. It tells us how often the classifier is correct for all positive instances. Low recall means that there is a high number of false negatives.

$$recall = \frac{TP}{TP + FN}$$

**F1-Score**: It is the harmonic mean of precision and recall.

$$F1\ score = 2 * \frac{precision * recall}{precision + recall}$$

**Confusion Matrix**: It is a table that is used to describe the performance of a classifier on the test data for which the true values are known (Table 1).

TABLE I. Confusion matrix

| | ***Predicted YES*** | ***Predicted NO*** |
|---|---|---|
| ***Actual YES*** | True Positives (TP) | False Negatives (FN) |
| ***Actual NO*** | False Positives (FP) | True Negatives (TN) |

## VI. Results

### A. *Machine learning models.*

Below are the results for the machine learning models – SVM and Naïve Bayes. The word embedding used to convert the features to vector is TF-IDF. The inputs for these models was the training data with Random under sampling applied to address the imbalance in the dataset.

TABLE II. Machine Learning Model Results – with under-sampling

| ***Model*** | ***Accuracy*** | ***Precision*** | ***Recall*** | ***F1 Score*** |
|---|---|---|---|---|
| ***SVM*** | 88.43 | 0.88 | 0.66 | 0.71 |
| ***Naïve Bayes*** | 77.43 | 0.85 | 0.60 | 0.60 |

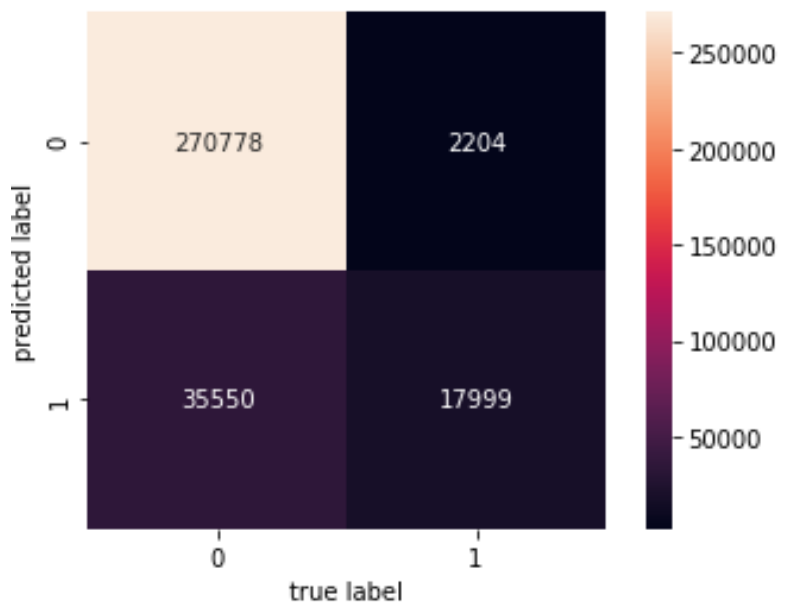

Fig. 6. Confusion Matrix – SVM – Undersampling – TF-IDF

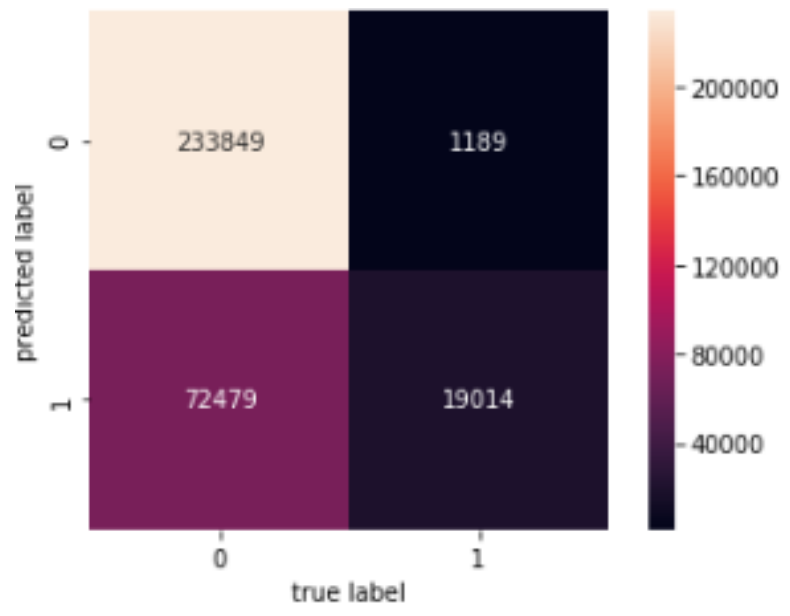

Fig. 7. Confusion Matrix – Naïve Bayes – Undersampling – TF-IDF

TABLE III. MACHINE LEARNING MODEL RESULTS – WITHOUT UNDER-SAMPLING

| *Model* | *Accuracy* | *Precision* | *Recall* | *F1 Score* |
|---|---|---|---|---|
| ***Naïve Bayes*** | 93.98 | 0.52 | 0.83 | 0.52 |

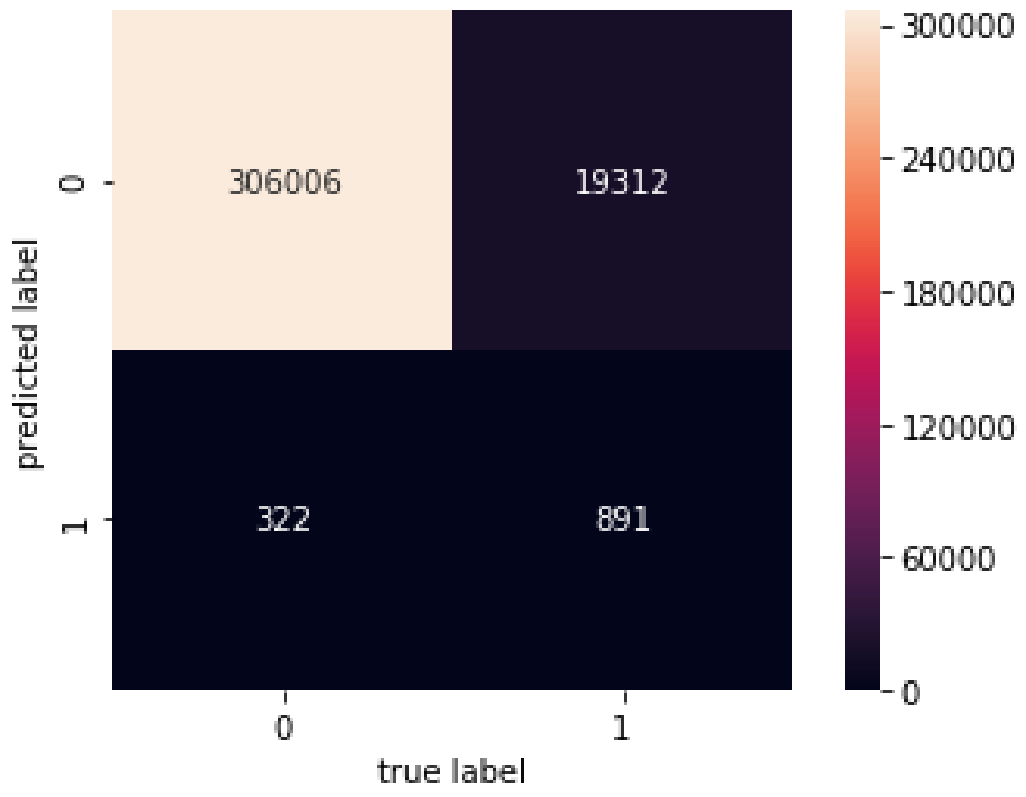

Fig. 8. Confusion Matrix – Naïve Bayes – No Undersampling – TF-IDF

SVM with under-sampling (approximately 50,000 samples in training data and 50,000 in test data) took very long time to train. This is because the fit time complexity for SVM is more than quadratic with the number of samples which makes it hard to scale to dataset with more than 10000-20000 samples. So we were not able to train SVM on the whole dataset.

Below are the results for the deep learning models – GRU. The word embeddings used to convert the features to vector are – without pre-trained embedding, using GloVe embedding, Using wiki-news fastText embedding. The inputs for these models was the training data with Random under sampling applied to address the imbalance in the dataset.

TABLE IV. RESULTS – CUDNNGRU- WITH UNDER-SAMPLING

| *Model CuDNNGRU* | *Accuracy* | *Precision* | *Recall* | *F1 Score* |
|---|---|---|---|---|
| ***Without pre-trained Embedding*** | 84.48 | 0.87 | 0.63 | 0.67 |
| ***Using Glove Embedding*** | 89.49 | 0.89 | 0.67 | 0.72 |
| ***Using Wiki news fastText Embedding*** | 85.84 | 0.88 | 0.64 | 0.68 |

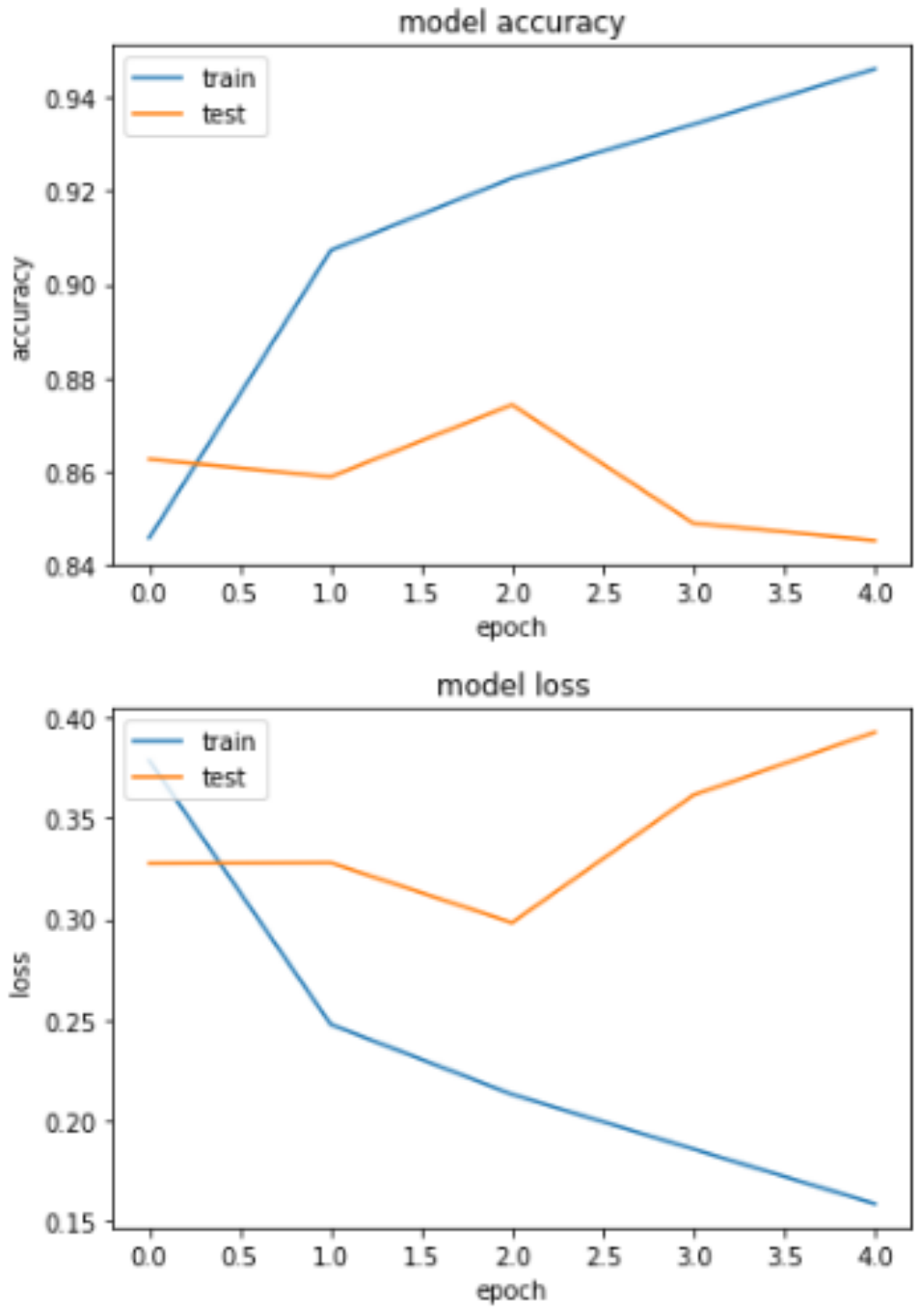

Fig. 9. Accuracy and Loss for CuDNNGRU – Undersampling - Without pre-trained Embedding

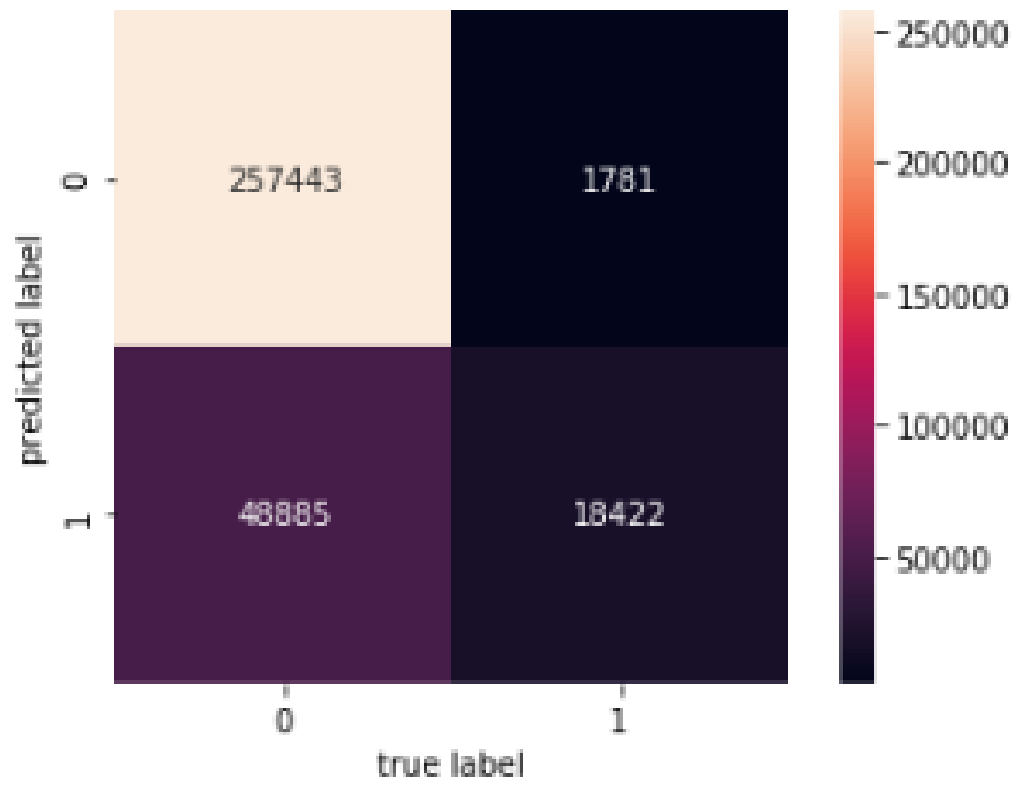

Fig. 10. Confusion Matrix – CuDNNGRU – Undersampling - Without pre-trained Embedding

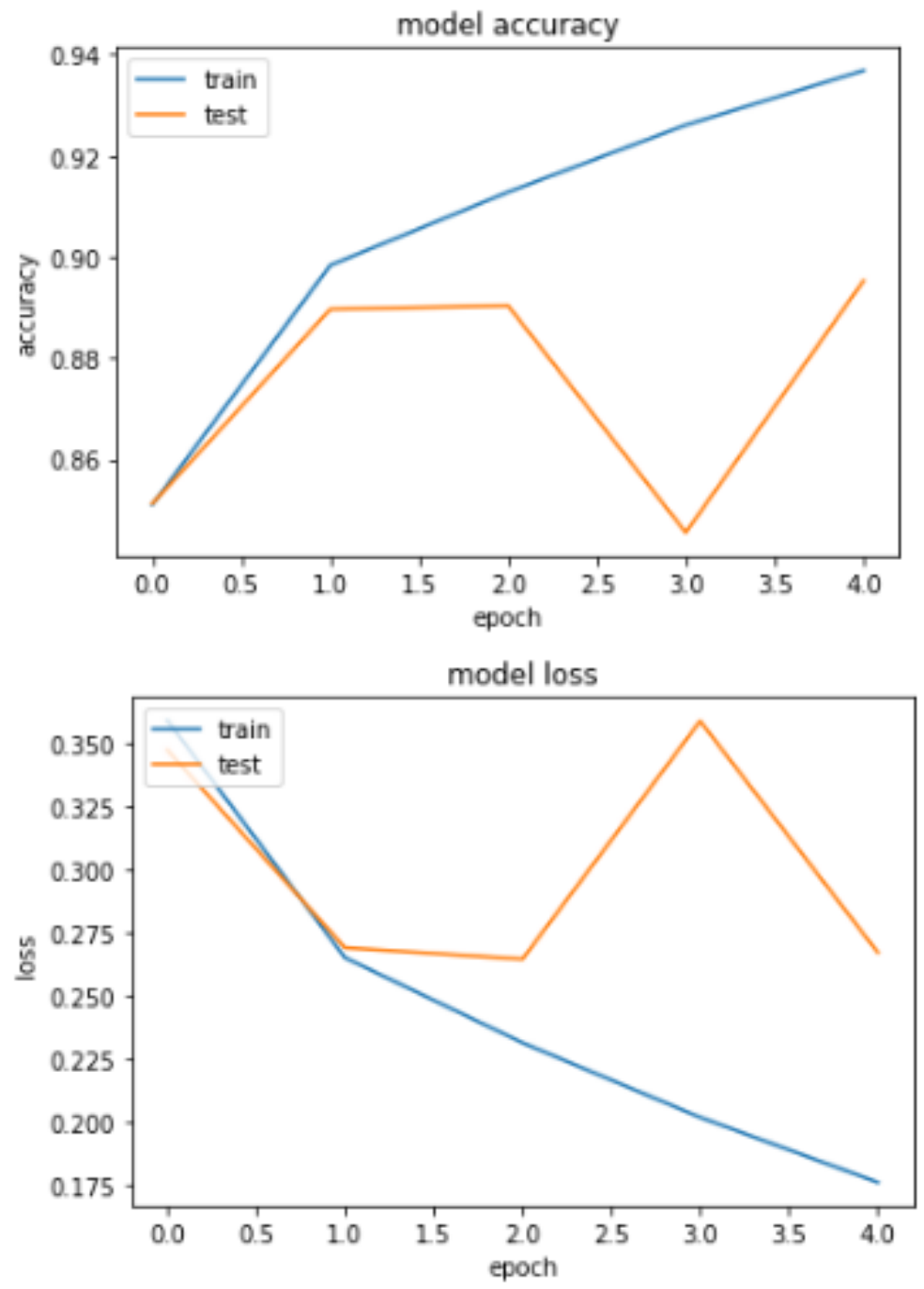

Fig. 11. Accuracy and Loss for CuDNNGRU – Undersampling - Using Glove Embedding

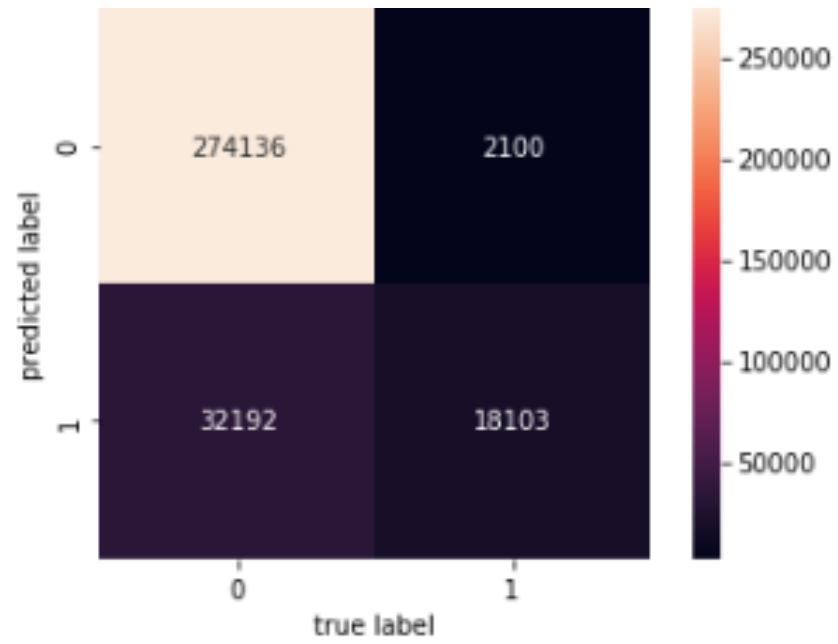

Fig. 12. Confusion Matrix – CuDNNGRU – Undersampling - Using Glove Embedding

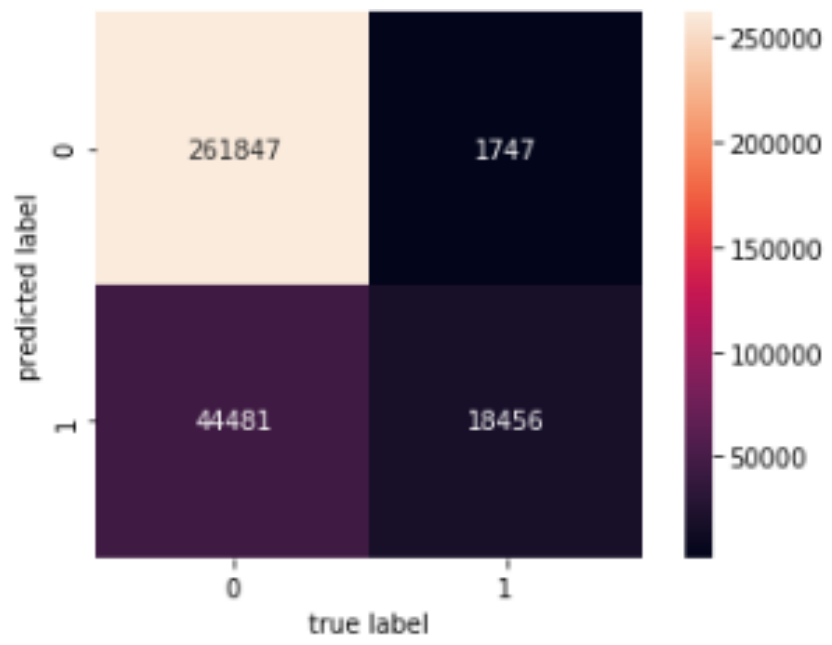

Fig. 13. Confusion Matrix – CuDNNGRU – Undersampling - Using Wiki News FastText Embeddings

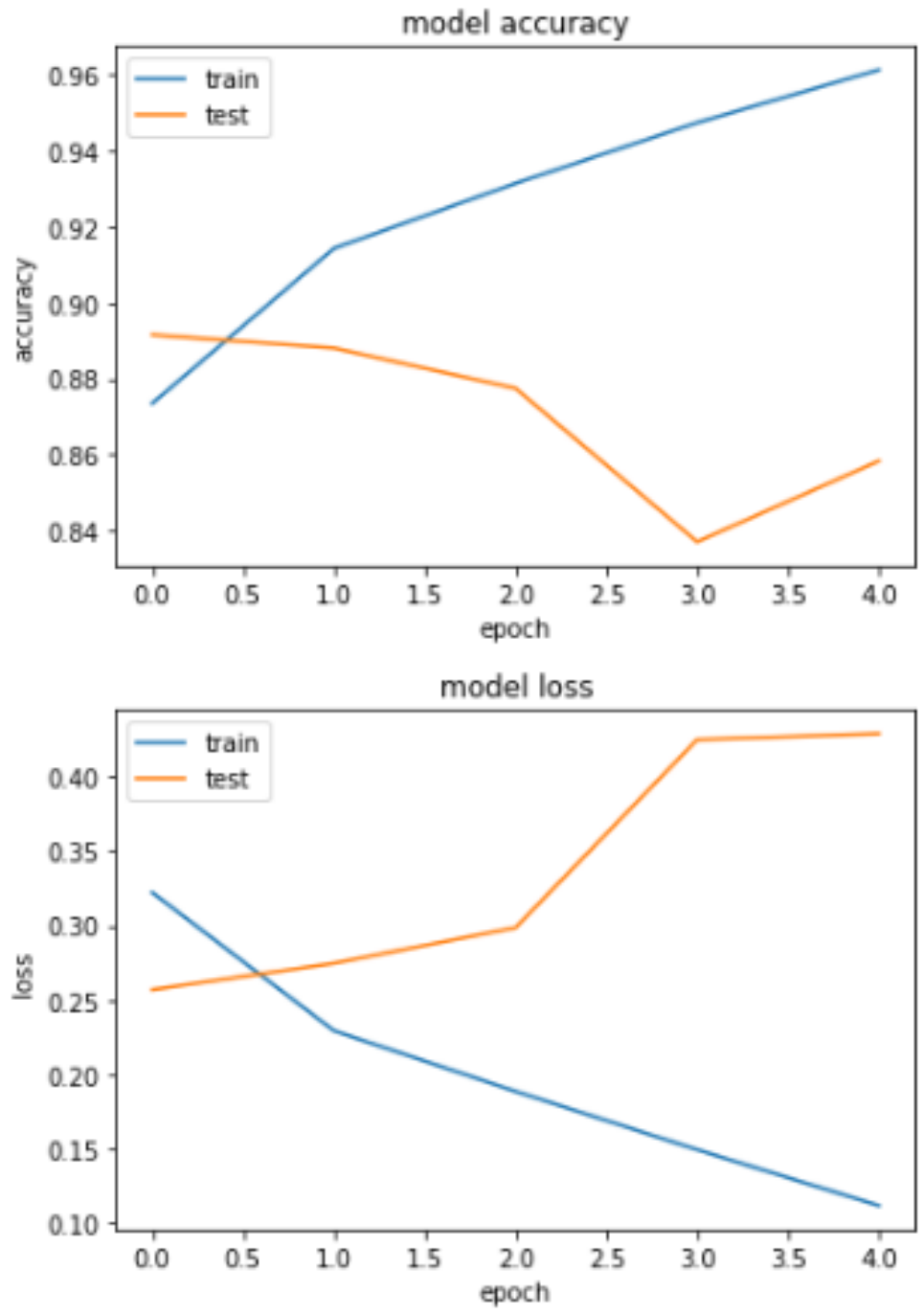

Fig. 14. Accuracy and Loss for CuDNNGRU – Undersampling - Using Wiki News FastText Embeddings

Below are the results for the deep learning models – LSTM. The word embeddings used to convert the features to vector are – without pre-trained embedding, using GloVe embedding, Using wiki-news fastText embedding. The inputs for these models was the training data with Random under sampling applied to address the imbalance in the dataset.

TABLE V. RESULTS – LSTM- WITH UNDER-SAMPLING

| *Model LSTM* | *Accuracy* | *Precision* | *Recall* | *F1 Score* |
|---|---|---|---|---|
| ***Without pre-trained Embedding*** | 87.79 | 0.88 | 0.65 | 0.70 |
| ***Using Glove Embedding*** | 87.53 | 0.89 | 0.65 | 0.70 |
| ***Using Wiki news fastText Embedding*** | 88.27 | 0.88 | 0.66 | 0.70 |

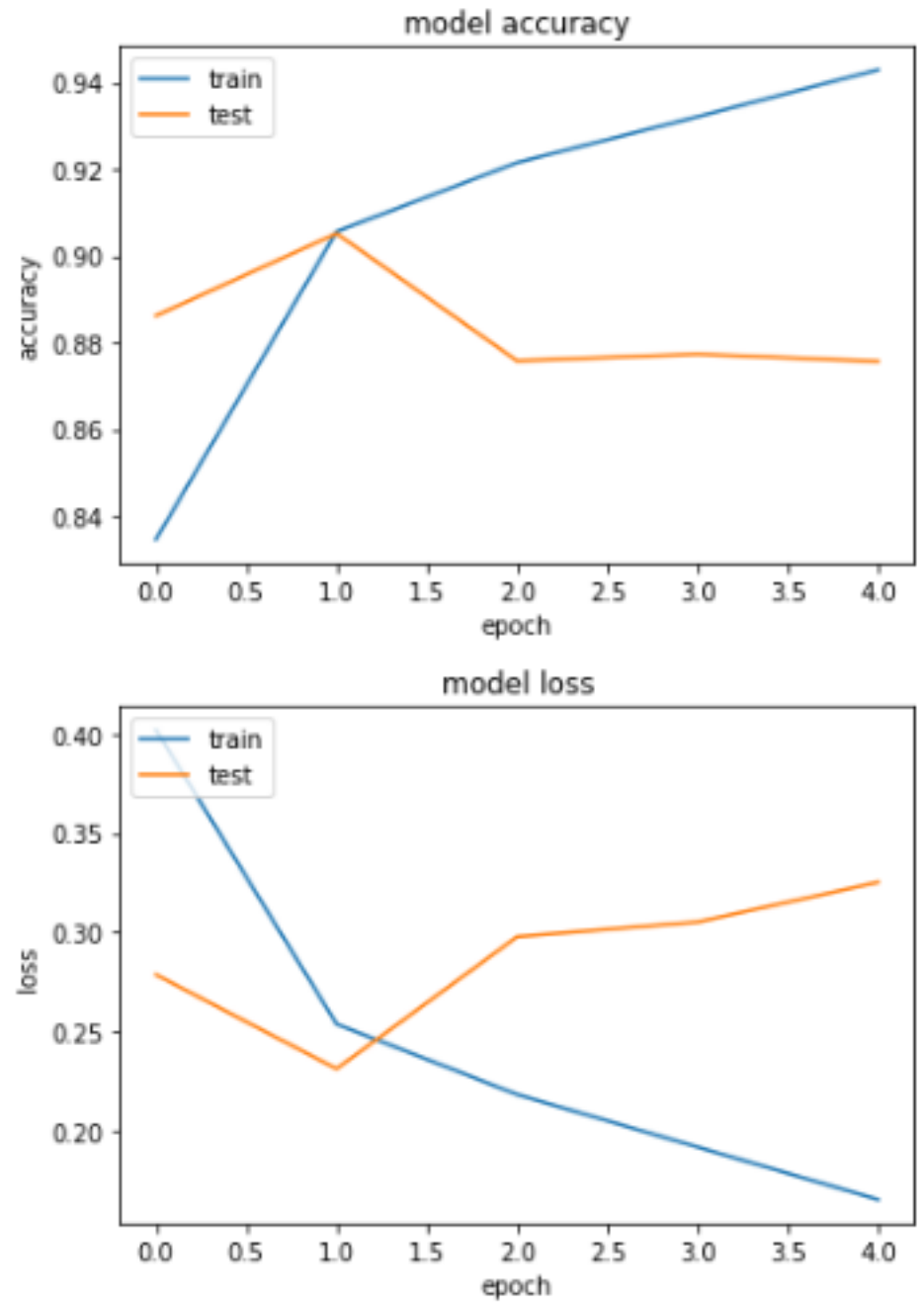

Fig. 15. Accuracy and Loss for LSTM – Undersampling - Without pre-trained Embedding

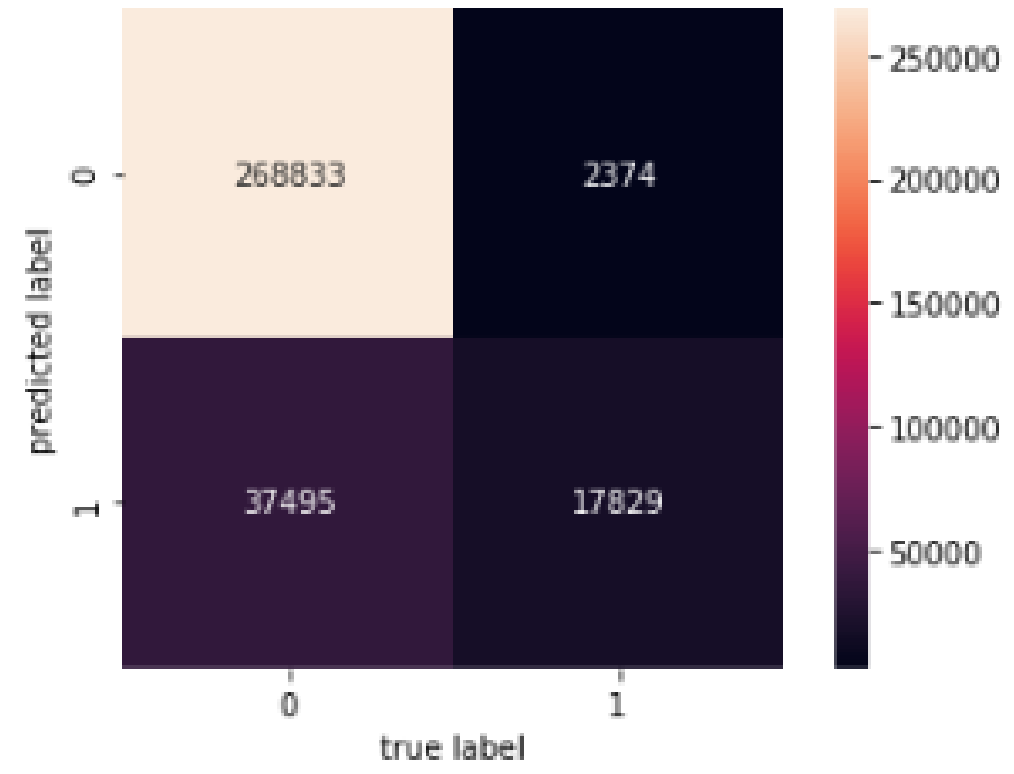

Fig. 16. Confusion Matrix – LSTM – Undersampling - Without pre-trained Embedding

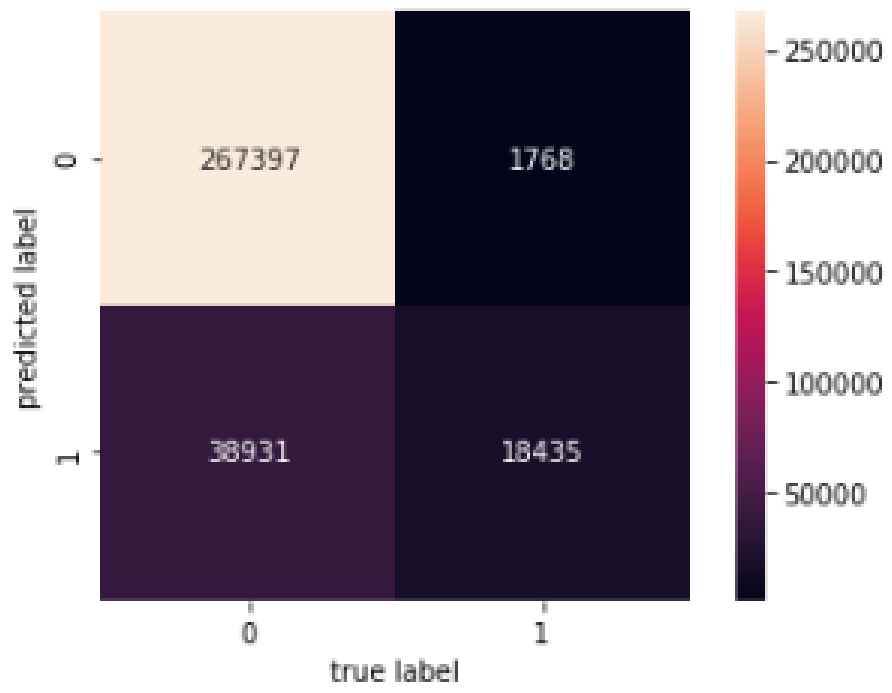

Fig. 17. Confusion Matrix – LSTM – Undersampling - Using Glove Embedding

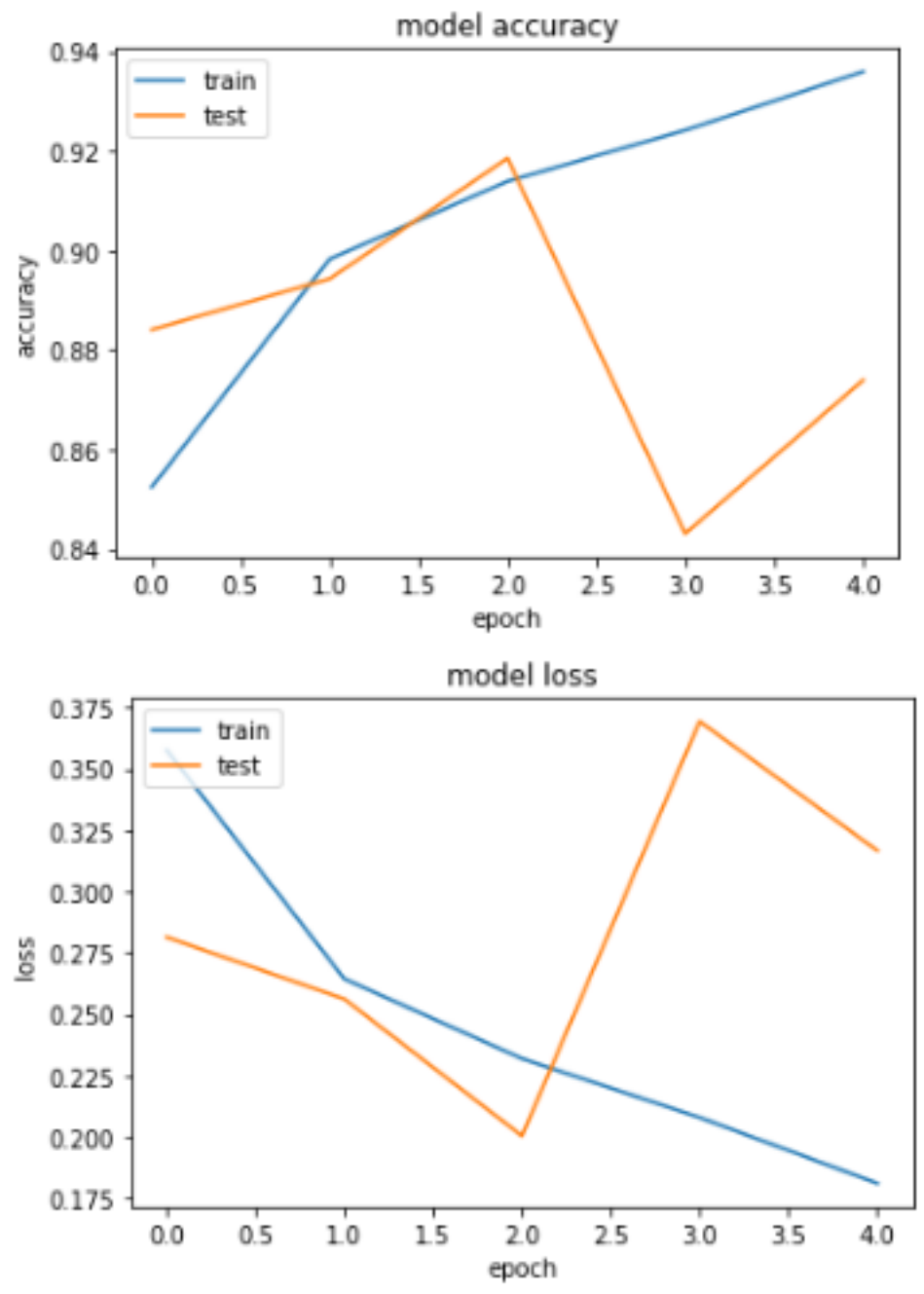

Fig. 18. Accuracy and Loss for LSTM – Undersampling - Using Glove Embedding

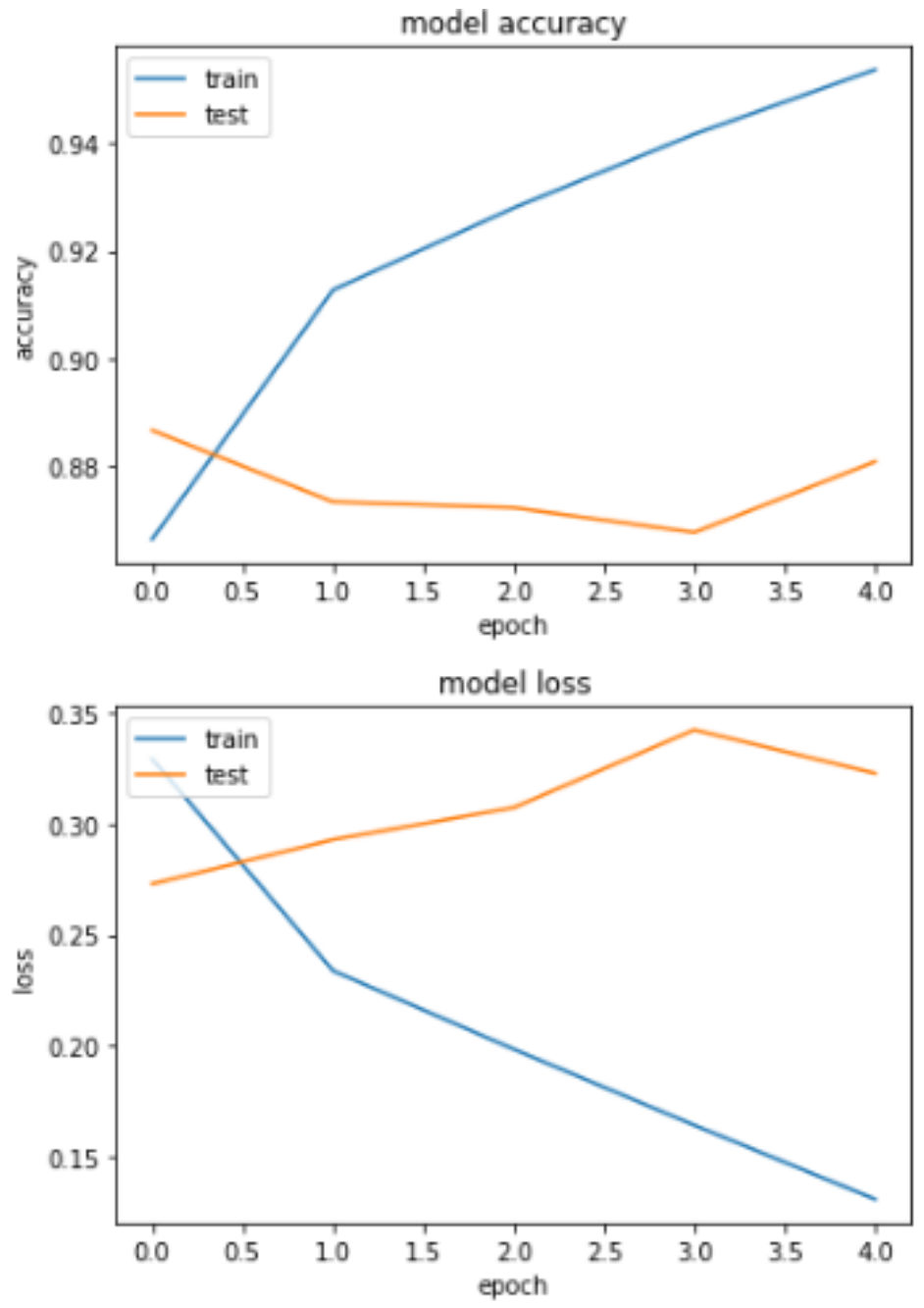

Fig. 19. Accuracy and Loss for LSTM – Undersampling - Using Wiki News FastText Embeddings

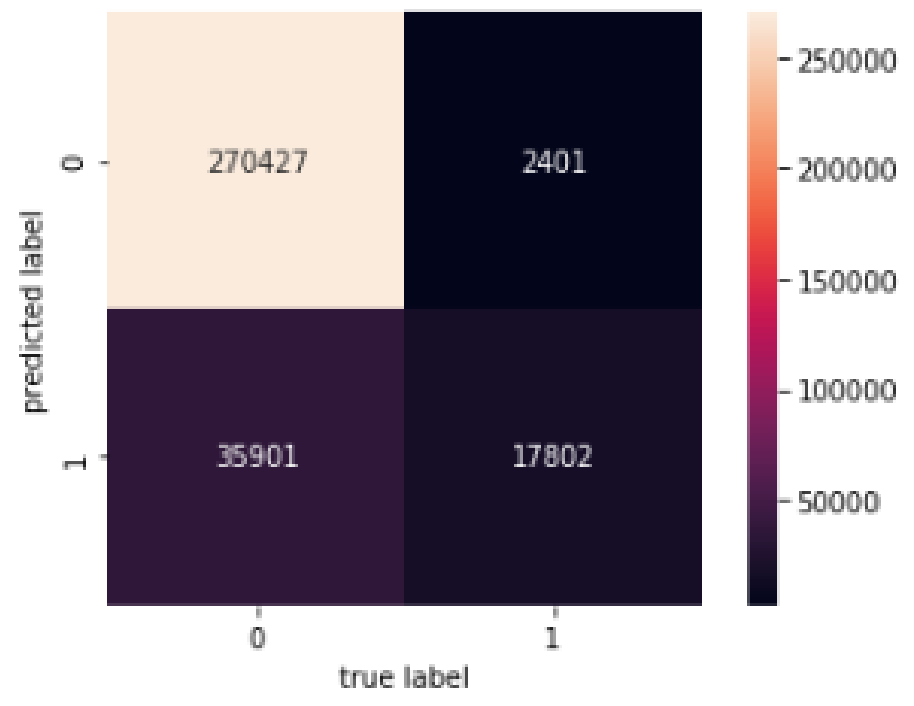

Fig. 20. Confusion Matrix – LSTM – Undersampling - Using Wiki News FastText Embeddings

TABLE VI. MISCLASSIFIED RESULTS – LSTM- WITH UNDER-SAMPLING

| *Model LSTM* | *Actual* | *Predicted* |
|---|---|---|
| ***How much time does it take to confirm an identity on Facebook?*** | 1 | 0 |
| ***Is GPA 2.56 less than GPA 2.59?*** | 1 | 0 |
| ***Why do men break promised the very next day?*** | 0 | 1 |
| ***Where can I find a beautiful, intelligent girl who isn't smug and conceited about it?*** | 0 | 1 |

We observed the predictions made by LSTM for some of the samples to compare the actual and predicted values. We note that, some of the samples that were marked as insincere in the dataset were actually sincere and the model has also classified them as sincere. Similarly, some samples that were marked as sincere in the dataset were actually insincere and the model has also classified them as insincere. This means that few errors made by LSTM were actually correct guesses marked incorrectly by noise in the dataset.

## VII. DISCUSSION

When comparing machine learning models, SVM with under-sampling (approximately 50,000 samples in training data and 50,000 in test data) took the longest time to train due to its high complexity. Naïve Bayes performed well on the data set both with and without under-sampling. However, under-sampling gave better accuracy for Naïve Bayes but the F1 score was really low, so in practice naïve Bayes with under-sampling should be used.

Among the deep learning models, GRU and LSTM, the performance of both the models were very much similar, probably because both have the ability to retain some data from the past. However, GRU gave slightly better results and the LSTM model was able to correctly predict some of the samples which are incorrectly classified in the dataset by humans.

We observe that the performances of the machine learning and deep learning models are comparable. Naïve Bayes takes the least time to train and predict the model with good accuracy.

Overall, GRU using GloVe embedding provided the best result ( Accuracy = 89.49, F1 score = 0.72)

## VIII. CONCLUSION

We were able to analyze the 'Quora insincere Questions Classification' dataset to classify the questions as sincere and insincere. We processed the dataset using various word embedding models such as GloVe, wiki-news and TF-IDF. We handled the dataset imbalance using Centroid based cluster and random sampler. We trained various models such as- SVM, Naïve Bayes, GRU and LSTM. We observed that Machine learning algorithms (SVM and Naïve Bayes) can achieve high accuracy for classifying insincere questions without handling the imbalance, i.e., they can learn from an imbalanced dataset. The best results were obtained using GRU when the imbalance was handled using random sampler and GloVe word embedding were used.

## IX. FUTURE WORK

The dataset provided is highly imbalanced, we could try to collect more samples for the insincere questions. The data also contains noise, questions that are not classified correctly by humans, some sort of screening could be applied to get rid of the noise. The dataset could be trained using the recent model BERT and its performance could be compared with that of GRU.